%% file: main.tex
\documentclass[times,twocolumn,final,authoryear]{elsarticle}

\usepackage{ycviu}
\usepackage{framed,multirow}
\usepackage{booktabs}

\usepackage{amssymb}
\usepackage{latexsym}

\usepackage{url}
\usepackage{xcolor}
\definecolor{newcolor}{rgb}{.8,.349,.1}

\newcommand{\UD}[1]{{\color{black}#1}}
\newcommand{\ud}[1]{\UD{#1}}

\usepackage{amsmath,amssymb} %
\usepackage[export]{adjustbox}
\usepackage{rotating,multirow}

\usepackage{algorithm}
\usepackage[noend]{algorithmic}
\usepackage{eqparbox}
\usepackage{sidecap}
\usepackage[labelfont=bf,textfont=bf]{caption}
\usepackage[labelfont=bf,textfont={bf}]{subcaption}

\usepackage{xspace}
\makeatletter
\DeclareRobustCommand\onedot{\futurelet\@let@token\@onedot}
\def\@onedot{\ifx\@let@token.\else.\null\fi\xspace}

\def\eg{\emph{e.g}\onedot} 
\def\ie{\emph{i.e}\onedot}

\makeatother

\def\-{\raisebox{.75pt}{-}}

\newcommand{\mt}[1]{\mathbf{#1}}

\newcommand{\myparagraph}[1]{\textbf{#1}\quad}

\newcommand{\D}{\Delta}
\newcommand{\E}{$\Delta$}

\journal{Computer Vision and Image Understanding}

\begin{document}

\begin{frontmatter}

\title{Automatic Generation of Dense Non-rigid Optical Flow}%

\author[1]{Hoàng-Ân \snm{Lê}\corref{cor1}} 
\cortext[cor1]{Corresponding author: 
Email: h.a.le@uva.nl;}
\author[2]{Tushar \snm{Nimbhorkar}}
\author[1,3]{Thomas \snm{Mensink}}
\author[1]{Anil S. \snm{Baslamisli}}
\author[1,2]{Sezer \snm{Karaoglu}}
\author[1,2]{Theo \snm{Gevers}}

\address[1]{Computer Vision Lab, University of Amsterdam}
\address[2]{3DUniversum, Amsterdam}
\address[3]{Google Research, Amsterdam}

\received{1 May 2013}
\finalform{10 May 2013}
\accepted{13 May 2013}
\availableonline{15 May 2013}
\communicated{S. Sarkar}

\input{sections/abstract}

\begin{keyword}
\KWD dataset \sep optical flow \sep non rigid

\end{keyword}

\end{frontmatter}

\vspace{-1em}
\input{sections/introduction.tex}

\vspace{-1em}
\input{sections/related_work.tex}
\vspace{-1em}
\input{sections/method.tex}
\vspace{-1em}
\input{sections/dataset.tex}

\vspace{-1em}

\input{sections/experiments.tex}

\input{sections/conclusion.tex}
\vspace{-1em}
\bibliographystyle{model2-names}
\bibliography{UvA-iccv19}

\end{document}

%% file: sections/abstract.tex
\begin{abstract}
There hardly exists any large-scale datasets with dense optical flow of non-rigid motion from real-world imagery as of today. \ud{ The reason lies mainly in the required setup to derive ground truth optical flows: a series of images with known camera poses along its trajectory, and an accurate 3D model from a textured scene. Human annotation is not only too tedious for large databases, it can simply hardly contribute to accurate optical flow.
}
To circumvent the need for manual annotation, we propose a framework to automatically generate optical flow from real-world videos. The method extracts and matches objects from video frames to compute initial constraints, and applies a deformation over the objects of interest to obtain dense optical flow fields. We propose several ways to augment the optical flow variations. 
Extensive experimental results show that training on our automatically generated optical flow outperforms methods that are trained on rigid synthetic data using FlowNet-S, LiteFlowNet, PWC-Net, \ud{and RAFT}. Datasets and implementation of our optical flow generation framework are released at~\url{https://github.com/lhoangan/arap_flow}.%
\end{abstract}

%% file: sections/introduction.tex
\section{Introduction}
\label{sec:Introduction}

Optical flow estimation has gained significant progress with the emergence of
convolutional neural networks (CNN). %
The first end-to-end architecture, FlowNet, is proposed by~\citet{Dosovitskiy2015}
and extended to FlowNet2 by~\citet{Ilg2016}, which achieves state-of-the-art results.
Notable improvements using domain knowledge and classical principles include LiteFlowNet~\citep{hui2018}, PWC-Net~\citep{Sun2018} and RAFT~\citep{Teed2020},
with 30, 17, and 60 times fewer parameters than FlowNet2, respectively.
Additionally, attempts for unsupervised learning are presented, such as
\citet{Meister2018} using occlusion-aware bidirectional flow estimation
and \citet{Liu2019SelFlow} learning by distilling reliable flow
estimations from non-occluded pixels. However, they are limited by the ability
to model the problem and contribution of component weights of the loss functions~\citep{Meister2018}.

With the design of CNNs for optical flow, there is a growing demand for large scale datasets with corresponding dense optical flow fields. 
However, large-scale datasets with real world imagery and known ground truths in terms of dense optical flow fields simply do not exist. 
The reason is that dense flow fields are neither measurable with a sensor nor trivial to annotate by humans.
\ud{
Optical flow annotation requires having a full matching of all points in the latent 3D space
for each image pair before projecting into the image space.
For example, to construct the KITTI datasets~\citep{Geiger2012,Menze2015}, point clouds from 10 consecutive frames are extracted and registered together and manually checked so that ambiguous or trivially wrong pairs are removed before being projected back to image space.
Such tasks require \emph{accurate} 3D positions and orientations of \emph{all} points in each image as well as a perfect matching algorithm, otherwise the flow fields will be sparse.
}
As a result, while being the largest optical flow dataset available today containing real world images, KITTI datasets contain only 200 pairs of frames with sparse flow fields. 
This is insufficient for supervised training of CNNs designed for optical flow estimation.

\input{sections/method_fig_pipeline.tex}

To resolve the data demand of CNNs, synthetic (generated)
data is often used. 
A well-known synthetic dataset of optical flow is MPI-Sintel~\citep{sintel12}, which uses images and annotations rendered from a computer-generated movie called Sintel. 
This dataset contains non-rigid optical flow and serves as a well-established basis for comparing CNNs.
However, for fully supervised training of CNNs, the number of frames in MPI-Sintel (around 2K) is still insufficient.

Large-scale synthetic datasets are often generated from CAD models, where objects are deformed by affine transformations (zooming, rotation, and translation) and projected on randomly transformed backgrounds.
This process is the basis of datasets like FlyingChairs~\citep{Dosovitskiy2015}.
Due to the large number of available frames, these datasets are useful for training optical flow CNNs. 
Yet, CNNs trained on rigid flow with repetitive textures fail to generalize well~\citep{Mayer2018}.

Training with non-rigid motion is important for optical flow in real world imagery as many real objects deform in a non-rigid manner. 
Unfortunately, non-rigid optical flow ground-truth is not available in the current datasets.%

In this paper, we present a new approach to automatically generate dense optical flow fields from real-world videos. 
As illustrated in Fig.~\ref{fig:full_pipeline}, our approach collects motion statistics from real-world videos by computing image correspondences between segmented objects-of-interest. 
Then, the segmented objects are warped to generate complex deformations according to physical principles to generate dense optical flow fields.
Our method generates large amounts of optical flow data consisting of natural textures and non-rigid motions, which can be used for training CNNs designed for optical flow estimations.

\noindent This paper has the following contributions:
\begin{itemize}
\setlength\itemsep{0mm}
    \item We introduce the first method to automatically generate dense non-rigid optical flow fields from real-videos;%
    \item We make available a dataset with 55K frames consisting of natural textures and non-rigid optical flow, created from the DAVIS~\citep{DAVIS2017} video dataset;  
    \item We extensively analyze optical flow network architectures trained using our generated optical flow fields.
\end{itemize}

The paper is organized as follows. We first review the current datasets being used
for training optical flow in Sec.~2. Then, Sec.~3 describes our approach to
generate a dense flow field from a pair of images. Afterwards, Sec.~4 explains how
the proposed approach is used to generate a dataset and explores the
characteristics of the dataset. Finally, Sec.~5 compares the efficiency of
our dataset with alternative approaches for optical flow training.

%% file: sections/method_fig_pipeline.tex
\begin{figure*}[t]
    \centering
    \includegraphics[width=.85\textwidth]{images/eccv20_diagram_tip2.png}
    \caption{Overview of the proposed pipeline to generate a dense optical flow field from two video frames. 
     (A) the objects of interest are extracted;
     (B) motion characteristic is captured by finding correspondences between the objects;
     (C) object deformation constrained by the correspondences results in a dense flow field;
     (D) the resulting flow field is used to warp the object; and
     (E) both the extracted first-frame object and the warped object are pasted on a random background.
    The resulting pair of frames is used to train a deep neural network with the obtained dense flow field as the ground truth.
    }
    \label{fig:full_pipeline}
\end{figure*}

%% file: sections/related_work.tex
\section{Related Work}
\label{sec:related_work}

Most of the benchmark datasets provide optical flow for synthetically generated scenes, including
 Sintel~\citep{sintel12}, 
FlyingChairs~\citep{Dosovitskiy2015}, 
and Body~Flow~\citep{Ranjan2018}.
Among them, only the KITTI datasets~\citep{Geiger2012,Menze2015} provide optical
flow for real-world images. However, these datasets are limited to around 200 frames of car-driving scenes and consist mostly of rigid motion patterns. 

\citet{Dosovitskiy2015} are the first to generate large-scale optical flow dataset. They warp 2D chair images rendered from CAD models using random affine transformations, hence the name FlyingChairs. The first frame of a pair is created by randomly positioning multiple chair images on an image background. Then, the second frame is generated by warping each object using a flow field generated by the affine model with random parameters. 
While the parametric model is able to generate many images, the affine transformation yields rigid optical flow fields limiting the type of motion. 

SlowFlow~\citep{Janai2017} exploits the linearity of small motions and tracks pixels through densely sampled space-time volumes using high-resolution and high-speed cameras ($>$1440p resolution and $>$200 fps). High spatial resolution provides fine texture details, while high temporal resolution ensures small displacements allowing to integrate strong temporal constraints. However, the requirement of using special recording devices  as well as the potential inaccuracies in the \textit{estimated} optical flow limits the applicability of the method.

\myparagraph{Data Augmentation.}
Data augmentation entails a plethora of strategies to create more training data.
Widely used techniques for augmenting image data is to perform
geometric augmentation (such as translation, rotation, and scaling) and color augmentation (such as changing brightness, contrast, gamma, and color). Data augmentation for optical flow networks is first proposed by~\citet{Dosovitskiy2015} and studied in detail by~\citet{Mayer2018}. The results show that both color and geometry types of augmentation are complementary and improve the performance. Inspired by these data augmentation techniques, we propose methods to increase the diversity of the generated optical flow data by texture augmentations.

In conclusion, large scale datasets with dense optical flow of non-rigid motion from real-world imagery are not available today. This is mainly due to the difficulty of human annotation to generate optical flow ground-truth. Instead, synthetic optical flow datasets with computer-generated imagery are widely preferred. 
To circumvent human annotation and the use of synthetic imagery data, we propose a framework to automatically generate dense non-rigid optical flow from real-world videos.

%% file: sections/method.tex
\section{Generating Image Pairs for Optical Flow}
\label{sec:Method}

In this section, we describe our approach to generate a dense optical flow field from a pair of images, see Fig.~\ref{fig:full_pipeline}.
The framework is described in the following sections as follows:
 {%
 \begin{enumerate}
     \itemsep1mm
     \item[3.1] \textit{Image segmentation} receives 2 image frames and extracts the object of interest from them;
     \item[3.2] \textit{Image matching} receives the extracted objects and  obtains corresponding points between them;
     \item[3.3] \textit{Image deformation} computes the flow fields, guided by the correspondences;
     \item[3.4] \textit{Warping} of the first object with the flow field to generate a warped object;
     \item[3.5] \textit{Random background} on which we paste the first object and the warped object as an input pair for training, with the optical flow field as ground truth.
 \end{enumerate}
 }

\subsection{Image Segmentation}
\label{sec:method_seg}
Our aim is to generate flow fields from non-rigid moving objects in videos. 
From a pair of sequential frames $\mathcal{I}_t$ and $\mathcal{I}_{t+\D}$ in 
a video sequence, where $\D$ is an offset between the frames ($\D=1$ indicates consecutive frames), 
the objects of interest $I_t$ and $I_{t+\D}$ are localized, see (Fig.~\ref{fig:full_pipeline}.A).

We compare different ways to localize the objects, including ground truth segments and by using a pre-trained Mask R-CNN~\citep{He2017}. 
Segments can be the entire image frame and do not need to correspond to the objects precisely (see Sec.~\ref{sec:shape_var} for more details).

To increase the amount of variations in object motion, different offsets between frames ($\Delta$) are explored (Sec.~\ref{sec:disp_var}). 
 The localization of objects is also used to replace textures while keeping their shapes (Sec.~\ref{sec:text_var}).

\subsection{Image Matching} 
\label{sec:method_match}
The generated flow fields should adhere to the non-rigid motion of the objects in videos.
To steer the computation of the  flow field, the statistics of the object motion are computed by finding image matches (or correspondences) between the segmented objects $I_t$, and $I_{t+\D}$, see Fig.~\ref{fig:full_pipeline}.B.

As the paper aims to generate datasets for training deep networks, the data-agnostic DeepMatching method by~\cite{Weinzaepfel2013} is used to obtain the motion statistics between image frames.
    The method is inspired by the hierarchical, multi-layer and correlational structure of deep convolutional networks, but it does not require training. The matching is performed in 2 stages: bottom-up correlation pyramid computation and top-down corresponding extraction. Analyses show that DeepMatching is robust to non-rigid deformations and repetitive textures (due to multi-scale correlation). The method is efficient in determining dense correspondences from strong image changes.

The algorithm obtains a set of point-to-point matches.
We denote with $\mathcal{M}\left(\mt{x}_t^{k}\right) = \mt{x}_{t+\D}^{k}$ the map of the pixel coordinates of the $k$-th pixel in $I_t$ to the corresponding pixel coordinate $\mt{x}_{t+\D}^{k}$ in $I_{t+\D}$. The obtained correspondences are quasi-dense and robust to non-rigid deformations and repetitive textures.

\subsection{Image Deformation}
\label{sec:method_arap}
To generate a dense flow-field, we deform the segmented object $I_t$ to match with $I_{t+\D}$, using the obtained image matches $\mathcal{M}$ to guide the deformation process,  see Fig.~\ref{fig:full_pipeline}.C.
The \emph{as-rigid-as-possible} (ARAP)~\citep{Alexa2000,Wang2008,Dvoroznak2014,devito2017opt}
principle is used to deform the objects. ARAP allows for large non-rigid deformations but also conforming to physical feasibility by minimizing scaling and shearing factors of the local image regions. 

The deformation method is illustrated in Fig.~\ref{fig:warpedflow_examples}.
We define a rectangular grid tightly bounding the object $I_t$, where each vertex corresponds to a pixel.
This grid is deformed (see Fig.~\ref{fig:warpedflow_examples}(b) to (c)) steered by image matches $\mathcal{M}$ and regularized by local deformations.
The dense flow field $\tilde{F}_{t\rightarrow{t+\D}}$ is obtained by interpolating the vertices before and after deformation.

Mathematically, the image deformation process is formulated as an energy optimization problem over the grid structure. We minimize per image the energy of a data fitting term weighed with a regularizer:

\begin{equation}
	\!\!E(\mt{d}, \mt{R}, \mt{x}_{t},\! \mathcal{M}) =\!\!  \sum_k\! w_\mathrm{fit} \  E_\mathrm{fit} (\mt{d}^k, \mt{x}^k_{t}, \mathcal{M}) + w_\mathrm{reg} \   E_\mathrm{reg}(\mt{d}^k, \mt{x}^k_{t},\mt{R}^{k}), 
	\label{eq:eng}
\end{equation}

\noindent where $\mt{d}$ denotes the deformed grid fitted to object $I_{t+\D}$. 
Following~\cite{Dvoroznak2014}, we set $w_\mathrm{fit}=10$ and $w_\mathrm{reg}=0.1$.
The data fit term is guided by the matches $\mathcal{M}$:
\begin{equation}
	E_\mathrm{fit} (\mt{d}^k, \mt{x}^k_{t}, \mathcal{M}) = \left\|\mt{d}^k\!-\!\mathcal{M}(\mt{x}_{t}^{k})\right\|^2.
\end{equation}
For pixel coordinates without an image match, $\mathcal{M}(\mt{x}_{t}^k)=\mt{x}_{t}^k$ is used instead.

As regularizer, the relative rigid rotation between neighboring pixels with an image wide rotation matrix $\mt{R}^{kj}\in\mathbb{R}^{2\times2}$ is used. 
This has been shown to enforce rigid rotation and translation~\citep{Wang2008}.
This yields:

\begin{equation}
	E_\mathrm{reg}(\mt{d}^k, \mt{x}^k_{t},\mt{R}^{k}) = \frac{1}{4} \sum_{j=1}^4{ \left\|\mt{R}^{kj}\left(\mt{x}_t^{kj}\!-\!\mt{x}_t^{k}\right)
    - \left(\mt{d}^{kj}\!-\!\mt{d}^{k}\right)\right\|^2},
\end{equation}

\noindent where $\mt{x}_t^{kj}$ denotes the $j$-th neighbor from pixel $k$ and $\mt{d}^{kj}$ the coordinates after deformation of $\mt{x}_t^{kj}$. 
The four connected pixels to each pixel are used as neighbors.

Eq.~\ref{eq:eng} is minimized with respect to $\mt{d}$ and $\mt{R}$, resulting in a non-linear least square problem, which is solved by the iterative Gauss-Newton method~\citep{devito2017opt}.

\input{sections/method_fig_examples.tex}

\subsection{Image Warping}
\label{sec:method_warp}
The dense optical flow field $\tilde{F}_{t\rightarrow{t+\D}}$ is obtained by bilinear interpolating $\mt{x}_t$ and $\mt{d}$.
Due to possible errors introduced by the matching and deformation algorithm, it is only an approximation of the true field $F_{t\rightarrow{t+\D}}$, and hence it does not necessarily transform the object $I_t$ to the exact shape of $I_{t+\D}$. 

To generate correct triples, image warping $\tilde{I}_{t+\D} = \mathcal{W}(I_t, \tilde{F}_{t\rightarrow{t+\D}})$ is used, see Fig.~\ref{fig:full_pipeline}.D.
This results in a correctly generated triple $(I_t, \tilde{I}_{t+\D}, \tilde{F}_{t\rightarrow{t+\D}})$%
\footnote{Image warping might introduce artifacts due to interpolation used, it is however commonly used, \eg in the FlyingChairs dataset~\citep{Dosovitskiy2015}}.
In Fig.~\ref{fig:warpedflow_examples}, segmented objects $I_t$ and $I_{t+\D}$ are illustrated with the obtained optical flow $\tilde{F}_{t\rightarrow t+\D}$ and the warped object $\tilde{I}_{t+\D}$, including some of the matching correspondence errors recovered by the warping process. 

\subsection{Background Generation}
\label{sec:method_bg}
To obtain a full frame image pair, the object $I_t$ and the warped object $\tilde{I}_{t+\D}$ are projected on a (static) background image (Fig.~\ref{fig:full_pipeline}.E).
This background image is randomly sampled from a set of 8K Public Domain images from Flickr of general scenery, similar to the approach used for the creation of the FlyingChairs dataset~\citep{Dosovitskiy2015}.
However, in contrast to the FlyingChairs dataset, static backgrounds are used instead of affine transformed backgrounds.

%% file: sections/method_fig_examples.tex
\begin{figure*}[t]
    \centering
    \includegraphics[width=.30\linewidth]{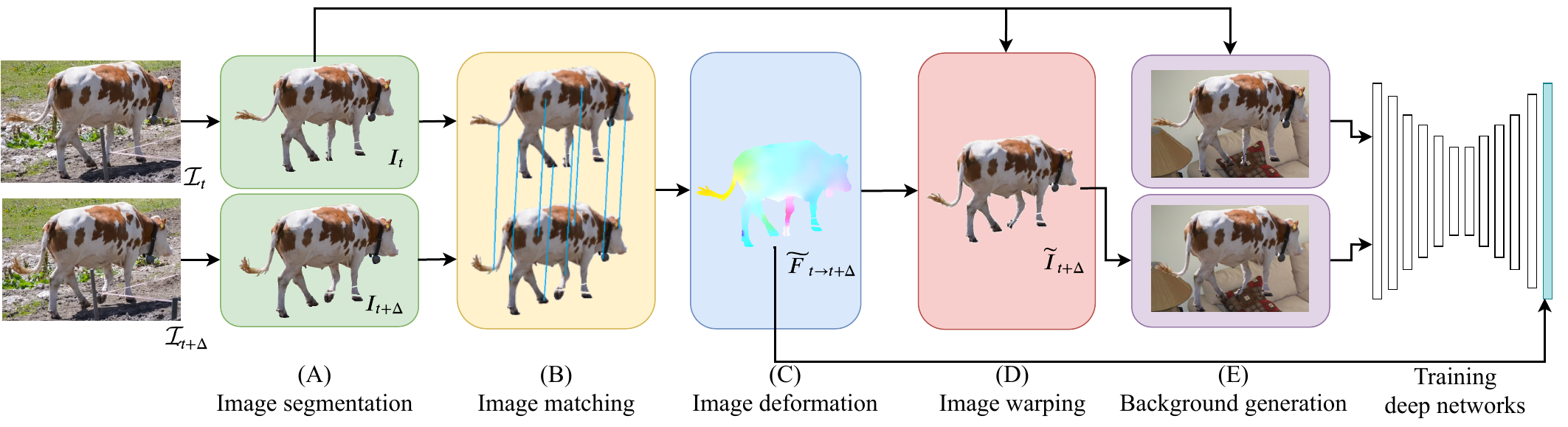}
    \includegraphics[trim={0 125mm 0 0},clip,width=.69\linewidth]{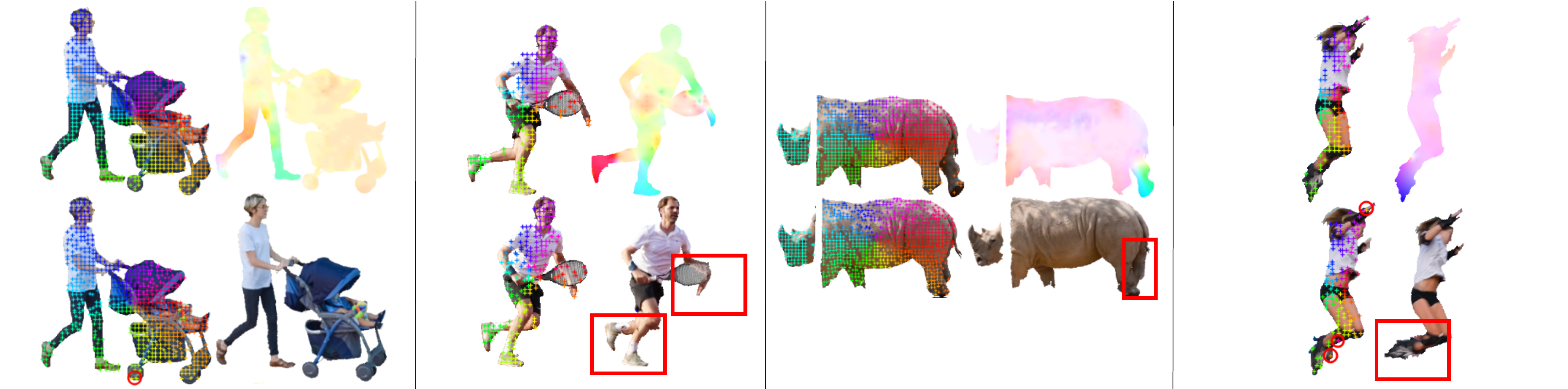}
    \caption{Illustration of ARAP image deformation. Left: deformation process includes constructing a control lattice (a), deforming the control lattice steered by the image matches (b,c), obtaining the flow field by interpolating the deformed lattice (d). Right: examples of image segments $I_t$ and $I_{t+\D}$ annotated with point matches, the computed flow $\tilde{F}_{t\rightarrow t+\D}$ and the warped images $\tilde{I}_{t+\D}$.
    Note the significant differences between  $I_{t+\D}$ and $\tilde{I}_{t+\D}$ (bottom row) due to the errors in the point matches. However $(I_t, \tilde{I}_{t+\D}, \tilde{F}_{t\rightarrow t+\D})$ form a correct triplet.
    }
    \label{fig:warpedflow_examples}
\end{figure*}

%% file: sections/dataset.tex
\section{Generating the DAVIS-Mask-OpticalFlow Dataset}
\label{sec:Variation}

In this section, datasets generated from video frames using the proposed method are explored. See the pseudo-code in Algorithm~\ref{alg:dataset}.
The algorithm takes a video dataset $\mathcal{V}$ as input, together with an integer number $\D$ for the frame distance, a segmentation method $S$, and a texture replacing method $T$. 
CNNs learn a better model when the training set consists of samples with a large variety of textures, motion patterns, and displacements~\citep{Mayer2018}.
Hence, different choices for $\D$, $S$, and $T$ to create optical flow datasets are investigated. 

\renewcommand{\algorithmicthen}{}
\begin{algorithm}[t]
     \caption{Generate optical flow from a video dataset $\mathcal{V}$.}
     \begin{algorithmic}[1]
         \renewcommand{\algorithmicrequire}{\textbf{Input:}}
         \renewcommand{\algorithmicensure}{\textbf{Output:}}
         \REQUIRE a video $\mathcal{V}$, frame-distance $\D$, segmentation method $S$, texture method $T$
         \ENSURE optical flow dataset $\mathcal{F}$         
         \FOR{$t \in \left[0, \text{length}(\mathcal{V}) - \D \right]$}
             \STATE $\mathcal{I}_t, \mathcal{I}_{t+\D} \gets$ sampled from $\mathcal{V}$%
            		\COMMENT {Sec.~\ref{sec:disp_var}}
    	\STATE $I_t, I_{t+\D} \gets$ segmenting $\mathcal{I}_t$ and $\mathcal{I}_{t+\D}$ using $S$
    		\COMMENT {Sec.~\ref{sec:shape_var}}	
    	\STATE $\mathcal{M} \gets$ image\_matching($I_t, I_{t+\D}$)
             \STATE \algorithmicif {$\ \ \mathcal{M} \ \textbf{is} \  \varnothing$: \textbf{skip frame}}
    	\STATE $\tilde{F}_{t\rightarrow{t+\D}} \gets$ ARAP deformation($\mathcal{M}$, $I_t$)
            \STATE $I_t \gets$ replace texture of object $I_t$ using $T$
            		\COMMENT {Sec.~\ref{sec:text_var}}
            \STATE $\tilde{I}_{t+\D} \gets$ image warping $\mathcal{W}(I_t, \tilde{F}_{t\rightarrow{t+\D}})$
            \STATE $\mathcal{I}_t, \tilde{\mathcal{I}}_{t+\D} \gets$ pasting $I_t$ and $\tilde{I}_{t+\D}$ on a random background
            \STATE $\mathcal{F}\gets\mathcal{F} + \{(\mathcal{I}_t, \tilde{\mathcal{I}}_{t+\D}, \tilde{F}_{t\rightarrow t+\D})\}$
        \ENDFOR
     \end{algorithmic}
     \label{alg:dataset}
 \end{algorithm}

The DAVIS~\citep{DAVIS2017} video dataset is used to generate optical flow datasets. 
DAVIS contains ~6K frames of real imagery with provided segmentation masks.
The generated optical flow datasets are used to train a FlowNet-S (FNS) model~\citep{Dosovitskiy2015}. 
Evaluation is performed on a subset of 410 image pairs from the training set of MPI-Sintel~\citep{sintel12}, coined Sintel-val.
Results are reported using the average \emph{end-point-error} metric (AEPE, lower is better).
The obtained results are compared to a FlowNet-S model trained on the FlyingChairs~\citep{Dosovitskiy2015} dataset as baseline. This dataset has 22K image pairs of chairs projected on different backgrounds with corresponding optical flow ground truths.
FlowNet-S is chosen since it is fast to train, thus suitable for extensive experimentation.
In Sec.~\ref{sec:Exp}, experiments are conducted with our final dataset using more recent architectures, and  a more diverse range of datasets.

\begin{figure}[t]
    \centering
    \begin{subfigure}[t]{0.54\columnwidth}
        \includegraphics[width=.8\textwidth, valign=b]{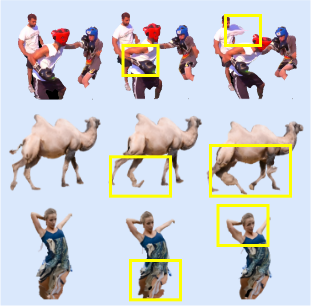}
        \caption{Warped images}
        \label{fig:displacement_warped}
    \end{subfigure}
    \begin{subtable}[t]{0.43\columnwidth}
    	\begin{adjustbox}{width=.9\textwidth,valign=b}
            \begin{tabular}{@{}lc@{}}\toprule
                 & Sintel-val \\
                \midrule
                FC    &   5.75  \\\midrule
                \E1 &   5.50 \\
                \E1-2         &   5.37 \\
                \E1-3       &   5.17 \\
                \E1-4     &   5.10 \\
                \E1-5     &   5.06   \\
                \E1-5,8     &   \bf 5.03   \\
                \E1-5,8,12     &   5.06  \\
                \bottomrule
            \end{tabular}
    	\end{adjustbox}
    	\caption{AEPE}    
	\label{tab:displacement_perf}
    \end{subtable}
    \caption{\textbf{Influence of the frame distance ($\D$)}:
    Increasing the frame distance increases the motion magnitude (a), and introduces artifacts in the warped objects (b), yet increasing the dataset by adding frame distances up to $\D=5$ is beneficial for performance(c). Larger frame distances have neglectable influence.
    }
\end{figure}

\subsection{Displacement Variation}
\label{sec:disp_var}
To increase the variation in object motion, different frame distances $\D$ in the video sequence are used.
Larger $\D$ reflects motion further in time, resulting in more variation in the non-rigid motion statistics of the objects, as shown in Fig.~\ref{fig:displacement_warped}.
Note, however, that larger frame distances also introduce artifacts, likely due to errors in the matching stage.

In order to study the influence of creating a dataset with larger frame distances,  
datasets generated with $\D=\{1, 2, \ldots, 12\}$ are combined into a single dataset.
Despite the increase of the training data size, the images' appearances basically stay the same as they are extracted from the same set of 6K videos. 
Hence, the performance gain is attributed to the increased displacement.

The performance is presented in the table of Fig.~\ref{tab:displacement_perf}.
From these results, it can be derived that increasing the frame distance is, in general, beneficial for AEPE error on Sintel-val.
We observe some diminishing gains, especially for $\D>5$. This is attributed to the introduced artifacts in the warped images.
More importantly, the results show that there is \emph{no} need to strictly match the distributions of the training and testing sets to achieve the best performance~\citep{Mayer2018}.
Although \E1-5,8 has slightly better performance, given that the large frame-distance version also produces more artifacts, this version is discouraged. The smaller set is also more favorable because of time efficiency. 
Thus, for the remaining experiments \E1-5 is used to generate optical flow, unless stated otherwise.

\begin{figure}[t]
    \centering
    \begin{subfigure}[t]{0.53\columnwidth}
        \includegraphics[width=\textwidth, valign=b]{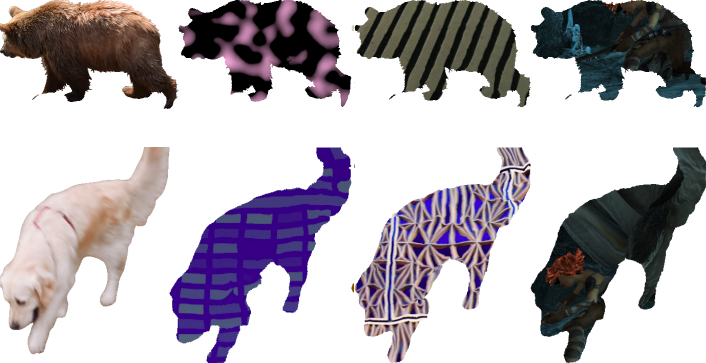}
        \caption{Appearance variation}
        \label{fig:texture_examples}
    \end{subfigure}
    \begin{subtable}[t]{0.46\columnwidth}
    	\begin{adjustbox}{width=\textwidth,valign=b}
            \setlength{\tabcolsep}{2pt} %
            \renewcommand{\arraystretch}{1} %
            \begin{tabular}{@{}lccccc@{}}\toprule
		&\multirow{2}{*}{T}  & \multicolumn{3}{c}{\E5} & \multirow{2}{*}{\shortstack[c]{Sintel\\val}} \\
		\cmidrule(lr){3-5}
             	& &    SynR & ReaR & SINv & \\
                \midrule
                FC &  &  7.29 & 6.92 & 6.28 & 5.10 \\
                \midrule
                \E1&O  & 4.82 & 4.62 & 4.20 & 5.50 \\
                \E1-4&O  & 4.75 & 4.42 & 4.02 & 5.10 \\
                \E1-4&R  & 4.61  & 4.32 & 3.99 & \bf 4.96 \\
                \E1-4&C  & \bf 4.60 & \bf 4.24 & \bf 3.86 & 4.98 \\
                \bottomrule
            \end{tabular}
        \end{adjustbox}
        \caption{Performance}
        \label{tab:texture_cross}
    \end{subtable}
    
    \caption{\textbf{Re-textured objects}: (a) examples of different textures, with left the original, and (b) performance analysis for different textures. 
    We conclude that training with re-textured objects (both R and C) ensures better performance.
    }
    \label{fig:texture}
\end{figure}

\subsection{Texture Variation}
\label{sec:text_var}

Object re-texturing allows for increasing the variation in the datasets appearances, see Fig.~\ref{fig:texture_examples}.
Moreover, re-textured objects enforce the network to disentangle semantic (class specific) information from optical flow information.
This is likely beneficial for a generic (class agnostic) optical flow prediction model. 

To re-texture objects, denoted by $T$ in Algorithm~\ref{alg:dataset}, the following procedure is followed.
After obtaining the flow field, the original texture (O) of $I_t$ is replaced by a new random texture (R), using the segmentation mask.
The random texture is taken from the set of general natural images utilized as background scenes.
Then, the corresponding $\tilde{I}_{t+\D}$ is obtained by warping the re-textured $I_t$, using $\tilde{F}_{t\rightarrow t+\D}$.
We explore using a re-textured dataset, denoted by R, and using a combination of original textures with random re-textures, denoted by C.

FlowNet-S is trained on the newly re-textured data ($\D1 \- 4$)  and its robustness for unseen textures and displacements is analyzed. 
The models are evaluated on \E5 and Sintel-val. The former has been re-textured with 3 texture types:
\emph{SynR}, synthetic images with repetitive patterns;
\emph{ReaR}, real images with repetitive patterns; and
\emph{SINv}, images from Sintel-val set, see examples of the re-textured images in Fig.~\ref{fig:texture_examples}.

The results are presented in the table of Fig.~\ref{tab:texture_cross}. We conclude that training with re-textured data (R or C) is beneficial for good performance on both \E5 and Sintel-val.
The performance differences between the different re-textured datasets used in \E5 show the dependency of the performance on the test images' texture.
This confirms the hypothesis that the network needs to be trained with a wide variety of texture types.
Hence, for the subsequent experiments, a combination of original-texture and re-textured images (C) is used.

\subsection{Object Segmentation}
\label{sec:shape_var}

\begin{figure}[t]
    \centering    
    \begin{subfigure}[t]{0.540\columnwidth}
        \centering
        \includegraphics[width=.7\textwidth, 
        valign=b]{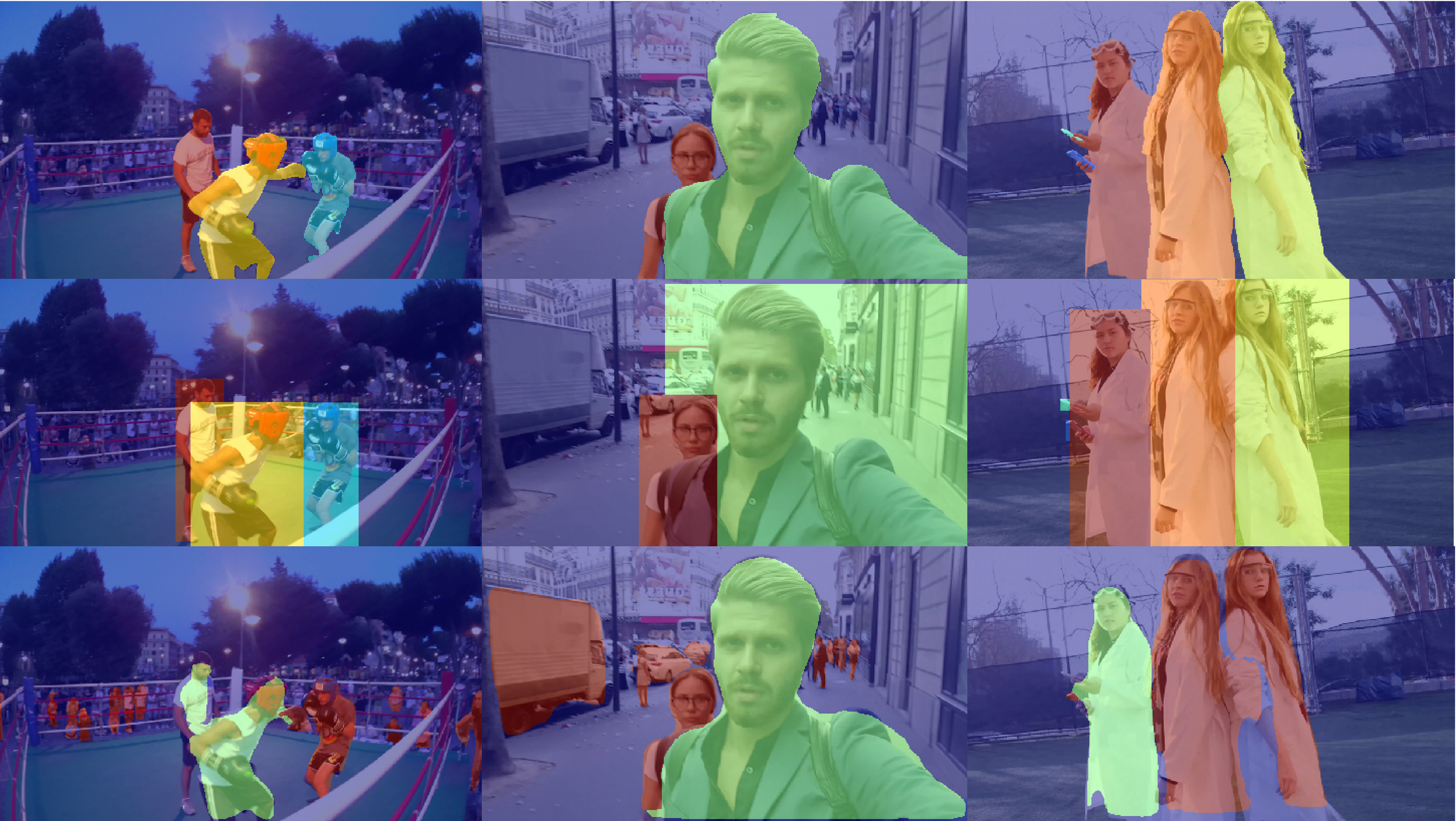}
        \caption{Segment variations}
        \label{fig:mrcnn}
    \end{subfigure}
    \begin{subtable}[t]{0.44\columnwidth}
        \centering
    	\begin{adjustbox}{width=.8\textwidth,valign=b}	
    	\setlength{\tabcolsep}{3pt}
            \begin{tabular}{@{}cc@{}}\toprule
                Segment type      & Sintel-val \\
                \midrule
                Full frame (F)    &   5.69 \\
                Ground truth (G)   &   5.06   \\
                Box  (B)  &   5.16   \\
                M-RCNN  (M)     &   \bf 4.91   \\
                \ud{SipMask  (S) }    &   5.09   \\
                \bottomrule
            \end{tabular}
        \end{adjustbox}
        \caption{Performance}
        \label{tab:maskrcnn}
    \end{subtable}
    \caption{\textbf{Object Segmentation}. (a) From top to bottom: examples of ground-truth segments (G), bounding boxes (B) and
    Mask R-CNN predictions (M); and (b) Sintel-val performance.
    The approximately correct segments M yield best performance. 
    (\E1-5-O)
    }
\end{figure}

In this section, different methods for selecting the object of interest are compared.
So far, the ground truth segmentation masks have been utilized. 
Now, the following alternatives are considered:
(1) selecting the entire frame as the object of interest;
(2) using tight bounding boxes enclosing the ground-truth segments;
(3) using the ground-truth segments; and
(4) using segments from Mask R-CNN~\citep{He2017} and \ud{SipMask~\citep{Cao2020_SipMask}}, pre-trained off-the-shelf segmentation networks.
See Fig.~\ref{fig:mrcnn} for examples.

Entire-frame deformations include constraints from both backgrounds and foreground objects, which might limit the flexibility and variation in the generated deformation.
The bounding boxes increase the segment sizes by including background parts while keeping the objects of interest in focus. 
For Mask R-CNN~\citep{He2017} and \ud{SipMask~\citep{Cao2020_SipMask}}, the available pre-trained models are used, which is trained on the class labels of the MS-COCO~\citep{Lin2014} datasets. 
Due to uncertainties in inference, the Mask R-CNN segments may generate larger regions rather than strictly focusing on the centred objects like the ground-truth  segments.
This might result in creating a large variation in terms of object shapes and sizes.
\ud{SipMask model is more accurate and can generate segments closer to the ground truth, thus reducing the variability of background segments which are shown to be useful and might reduce the network performance. The performance on Sintel-val is indeed close to that of using GT segments.}

The results of training FlowNet-S on the data generated using the original textures (E1-5-O), comparing different segmentation methods, are provided in the table of Fig.~\ref{tab:maskrcnn}.
It can be concluded that focusing on objects is beneficial with the performance being $F < B < G$. 
Surprisingly, the network trained with the dataset using Mask R-CNN segments outperforms the ones using ground truth segments and \ud{SipMask} ($M > G > S$). 
This is because Mask R-CNN segments, in general, have more variation and cover more object types in a scene compared against the ground truth \ud{and SipMask} segments: 
not only the objects of interest, but also those in the background. 
Hence, it provides the network with a larger range of patterns, which appears to be useful for training.
This indicates that it is possible to use any real-world in-the-wild videos with Mask R-CNN segments for training optical flow deep networks.
In the subsequent experiments, datasets using Mask R-CNN (M) are utilized.

\subsection{Non-Rigid Motion Analysis}
The Sintel movie is created using mostly static scenes (backgrounds) and moving characters (objects). 
In this section, the performance of the FNS models on non-rigid movements and occluded regions is evaluated.
Evaluation is performed over 
(a) foreground objects using the segments provided by~\citep{sintel12} and
(b) occluded regions, defined as those regions which appear in one of the two input frames only. 

\begin{table}[t]
	\centering
    \caption{\textbf{Non-Rigid Motion Analysis}: Performance on subsets of Sintel-val using the full image (F), non-rigid motion (N-R) or occluded regions (Occ), for three different architectures: FlowNet-S (FNS), PWC-Net (PWC), and LiteFlowNet (LFN). 
    Deep networks trained on our non-rigid motion datasets outperform those trained on the FlyingChairs dataset. 
    }
	\begin{adjustbox}{width=\linewidth,valign=b}
    \setlength{\tabcolsep}{2pt}
    \begin{tabular}{@{}cccccccccc@{}}\toprule
            & \multicolumn{3}{c}{FNS}
            & \multicolumn{3}{c}{LFN}
            & \multicolumn{3}{c}{PWC} \\
    \cmidrule(l){2-4}
    \cmidrule(lr){5-7}
    \cmidrule(r){8-10}
            & Full & N-R & Occ
            & Full & N-R & Occ
            & Full & N-R & Occ \\
        \midrule %
	FC      & 5.09 & 14.56 & 12.32 &
		4.32  &  14.70 & 12.86 &
        4.01  &  13.52 & 11.27 \\ 
        \E1-5-O-M    & 4.96 & 13.88 & 12.64 &
        4.34   &  \bf 13.59 & 12.88  &
                3.88 &   12.59 & 11.21 \\
        \E1-5-C-M    &  \bf 4.54 &  \bf 13.28 & \bf 11.82 &
        \bf 4.23 &  13.83 & \bf 12.79 &
        \bf 3.62 &  \bf 11.90 & \bf 10.57 \\
        \bottomrule
    \end{tabular}
    \end{adjustbox}
  \label{tab:conclu_var}	
\end{table}

\begin{figure}[t]
    \centering    
    \includegraphics[width=.8\linewidth]{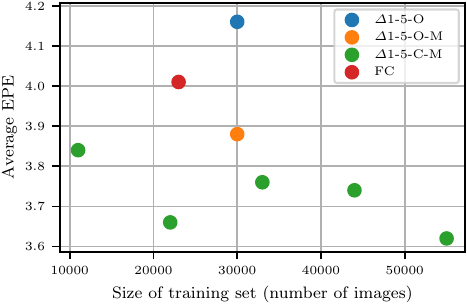}
    \caption{\textbf{Size of Training Set}: Performance on Sintel-val of PWC-Net~\citep{Sun2018} trained on differently sized datasets. 
    Results of sub-sampled \E1-5-C-M are indicated in green. 
    PWC-Net trained on \E1-5-C-M outperform the others, regardless of training size.
    }    
    \label{fig:traingsize}
\end{figure}

Different training sets are compared (FC, \E1-5-O-M, and \E1-5-R-M) using FlowNet-S~\citep{Dosovitskiy2015}, PWC-Net~\citep{Sun2018}, and LiteFlowNet~\citep{hui2018}.
The results are provided in Table~\ref{tab:conclu_var}.
We conclude that training on our non-rigid motion datasets outperforms training on the rigid transformations from the FlyingChairs dataset.
This holds especially for non-rigid and occluded regions in the images. 
All in all, non-rigid motion is a necessity to train robust CNNs for optical flow prediction. 

\subsection{Training Dataset Size}
In this section, the performance is evaluated as a function of number of images in the training dataset.
The aim is to distinguish improvements based on the sheer number of annotated training examples from the improvements based on non-rigid motion and texture variations.
Since the FlyingChairs dataset contains $\sim$22K images, the 55K images from our \E1-5-C-M dataset are sub-sampled to generate training sets with 11K, 22K, $\ldots$, 55K examples.

PWC-Net is trained on these different datasets and evaluated on Sintel-val.
The results are provided in Fig.~\ref{fig:traingsize}. 
It can be observed that the training set size does have an influence on the performance.
The results show that \E1-5-C-M gradually improves as the size of datasets increases (with 22K being the outlier likely due to sub-sampling effects).
However, the results also show that regardless of the size of this dataset, training on \E1-5-C-M performs the best.

\subsection{Discussion}
From these extensive yet initial analyses of various design choices, we derive that the generated datasets with non-rigid optical flow fields are well suited for training CNNs for optical flow prediction.
In the next section, the \E1-5-C-M dataset, coined Deepmatching-Maskrcnn-OpticalFlow (\textbf{DMO}), generated using Mask R-CNN segments using original and re-textured objects is utilized.

%% file: sections/experiments.tex
\section{Comparison using state-of-the-art architectures}
\label{sec:Exp}
In this section, experiments are conducted to compare models trained on the proposed DMO dataset to various state-of-the-art (SOTA) baselines and benchmarks.

\subsection{Experimental Setup}
\myparagraph{Datasets}
For most of the experiments, models trained on the FlyingChairs dataset~\citep{Dosovitskiy2015} are used to provide baseline comparisons.
This allows to analyze the effect of the training set on the performance of different network architectures.

Evaluation is performed on the test-set of MPI-Sintel~\citep{sintel12} containing large displacements of non-rigid optical flow.
Additional evaluations are performed on the validation split of the HumanFlow~\citep{Ranjan2018}, which contains 530 image pairs of non-rigid motion of human bodies, and on a subset of 50 randomly selected images from the KITTI 2012~\citep{Geiger2012} and KITTI 2015~\citep{Menze2015} training set containing real-textures from a self driving car.
Both MPI-Sintel and HumanFlow contain CGI-rendered images, and KITTI contains mostly rigid motion. The quantitative results on real-world non-rigid images can be found in the supplementary materials.%
Additionally, the QUVA repetition dataset~\citep{runia2018} is used to qualitatively evaluate on real videos with non-rigid motion.
It consists of video sequences of repetitive activities recorded in real life with 
minor camera motion, mostly consisting of non-rigid object motions.

\myparagraph{Network Architectures}
Different CNN architectures are compared, namely: (i) 
FlowNet-S~\citep{Dosovitskiy2015} which is also used in Sec~\ref{sec:Variation},
(ii) LiteFlowNet~\citep{hui2018}, PWC-Net~\citep{Sun2018}, \ud{and RAFT~\citep{Teed2020}} as recent supervised models, (iii) three unsupervised architectures, specifically MFOF~\citep{Janai2018ECCV}, DDFlow~\citep{Liu2019DDFlow}, and SelFlow~\citep{Liu2019SelFlow}. For each model, the standard training settings are used, including data augmentation and learning schemes as provided by the authors.

\ud{

\subsection{Performance with SOTA matching and segmentation}
\label{sec:raft}

\begin{table}[t]
    \centering
    \caption{Performance of RAFT being trained on datasets generated with different matching and segmentation algorithms. The best results of the entire column are shown in boldface while those among ARAP-based methods are underlined. The dataset generated by DeepMatching and Mask-RCNN, i.e. DMO, shows balancing performance between final and clean pass.}
    \begin{adjustbox}{width=.7\columnwidth,valign=b}
    \begin{tabular}{@{}cccccc@{}}\toprule
 	\multicolumn{2}{c}{\multirow{2}{*}{Training dataset}} & \multicolumn{2}{c}{Sintel-val} & \multicolumn{2}{c}{Sintel-test} \\
 	\cmidrule(r){3-4}\cmidrule(r){5-6}
        & & final & clean & final & clean \\
        \midrule        
        \multicolumn{2}{c}{FlyingChairs} & 4.25 & \bf 2.91 & 7.69 & \bf 4.48 \\
        DM & M-RCNN & 3.98 & \underline{3.12}& \bf \underline{6.15}& \underline{4.86}\\
        DM & SipMask & 3.91 & 3.27 & 6.20 & 4.88 \\
        NCNet & M-RCNN & 3.92 & 3.30 & 6.44 & 5.48 \\
        NCNet & SipMask & \bf \underline{3.81}& 3.45 & 6.57 & 6.55 \\
        \bottomrule
    \end{tabular}
    \end{adjustbox}
    \label{tab:raft}
\end{table}    

In this experiment, the datasets generated using different matching and
segmentation methods are compared by training the state-of-the-art flow prediction
network RAFT~\citep{Teed2020}. In particular, we compare the use of
DeepMatching (DM)~\citep{Weinzaepfel2013} in the previous experiments with
NCNet~\citep{Rocco2018_ncnet} for matching and Mask-RCNN~\citep{He2017}
with SipMask~\citep{Cao2020_SipMask} for segmentation. The results are presented in
Table~\ref{tab:raft}, where the combination (DM, M-RCNN) corresponds to the DMO dataset.
The performance evaluations from training with the FlyingChairs dataset are also included.

RAFT being trained on the FlyingChairs dataset achieves superior performance on the
clean pass of the Sintel dataset (both Sintel validation and test set), while being
inferior by a large margin on the final pass. This is  somewhat surprising as the final pass
is considered more difficult because of the severe environmental and motion
artifacts~\citep{sintel12}. From the training data viewpoint, it could be explained that the networks trained with ARAP-based datasets are more robust to artifacts as the training data contain artifacts from the matching and deformation processes
(Fig.~\ref{fig:warpedflow_examples}), while
the FlyingChairs 
images are all clean-cut (due to well-defined rigid transformation).

Among those using the same matching method, datasets using Mask-RCNN help the network produce slightly
higher performance. This is consistent with the results of the table in Figure~\ref{tab:maskrcnn}
which shows that the high accuracy of SipMask segmentation focusing mostly on image central objects (similar
to ground truth segmentation) are undesired for optical flow datasets.
On the other hand, among those using the same segmentation, the networks trained with datasets using
DeepMatching perform better. This is because NCNet is prone to repetitive patterns, especially when there are large changes in scale and locally geometrically consistent groups
of incorrect matches~\citep{Rocco2018_ncnet}, while DeepMatching is designed to 
be robust to repetitive patterns~\citep{Weinzaepfel2013}.

Considering the performance balance in both final and clean pass, we confirm the usage of
the DMO dataset, i.e. generated by DeepMatching and Mask-RCNN, in the following experiments, unless stated otherwise.

}

\input{sections/experiments_table_sota.tex}

\subsection{Performance with SOTA flow prediction networks}
We compare different state-of-the-art algorithms for optical flow, namely 
LiteFlowNet (LFN)~\citep{hui2018}, PWC-Net~\citep{Sun2018}, \ud{and RAFT~\citep{Teed2020}}. For each network, we compare the performance between the models trained with DMO dataset against the same model trained on FlyingChairs. The results are evaluated on the MPI-Sintel benchmark server (Sintel-test), the HumanFlow and KITTI 2012, 2015 datasets in Table~\ref{tab:sota}.

Except for RAFT, the networks trained with our dataset outperform those trained with FlyingChairs
on most of the cases. The results on Sintel occluded regions show that the proposed dataset
improves models' robustness even in the challenging occlusion conditions.
The results on HumanFlow show that 
non-rigid optical flow estimation (e.g. human body movement) benefits from DMO. 
In particular, although FNS is well-known for poor performance on 
small-displacement data~\citep{Ranjan2018}, the performance on HumanFlow trained
with DMO is close to that of the other powerful algorithms,
whereas FNS trained on FlyingChairs is close to zero-flow, which is consistent with the conclusion of \cite{Ranjan2018} as FlyingChairs was used in their experiment. This suggests that
the weakness of the methods can be improved using our proposed dataset.
\ud{ 
The results are mixed for RAFT and consistent to the results in table of Fig.~\ref{tab:maskrcnn} where
FlyingChairs more superior on Sintel clean pass and HumanFlow as they are both clean-cut and contain no artifacts. On the other hand, RAFT trained with the artifacts of DMO are more robust to the artifacts in the Sintel final pass.
This is an interesting benefit of artifacts that could be further exploited to improve optical flow datasets.

It should also be noted that both Sintel and HumanFlow data are computer generated (synthetic), while DMO contains only real-world textured images.
}
Moreover, the results on KITTI val set indicate how the networks perform on real-world textured images. The network trained with our dataset outperforms those trained with FlyingChairs. The performance gaps are larger with more sophisticated networks, such as PWC and RAFT. This shows the importance to have real-world textured datasets for real-world scenarios.

We conclude that optical flow CNNs benefit from training on natural textures and non-rigid movements, as generated by our dense optical flow method.

\input{sections/experiments_table_ft.tex}
\subsection{Performance on unsupervised and finetuned methods}
In this section, we measure the ability of our dataset DMO to transfer to different
domains. To this end, we compare PWC trained on our dataset
with unsupervised methods such as MFOF~\citep{Janai2018ECCV}, DDFlow~\citep{Liu2019DDFlow}, and SelFlow~\citep{Liu2019SelFlow}. We train
the network on a dataset generated in an unsupervised way, without using
any ground truth optical flow from the test domain, i.e. the MPI-Sintel
dataset.
We also show the results of finetuning on the Sintel train set with results reported by finetuned state-of-the-art methods.
The results are provided in Table~\ref{tab:unsupervised}.

For the transferred methods, PWC trained on our dataset
outperforms the others, showing the transferrability of our method. For the finetuning
methods, SelFlow performs the best. 
Note the improvement of the supervised results (Ft) over the unsupervised (U) and how it is trained on Sintel-related data ( pretrained on SintelMovie, finetuned with MPI-Sintel flow ground truth). 
This shows the necessity of having annotated data on target domains.

\input{sections/experiments_figure_quva.tex}

\subsection{Performance on real-world images}
As there are currently no optical flow benchmarks with real-texture and
non-rigid motion, we qualitatively show the results of PWC-Net and LiteFlowNet
on real world images.
Figure~\ref{fig:quva} presents the optical flow prediction by LiteFlowNet (top)
and PWC-Net (bottom) trained with FlyingChairs and our DMO
on the QUVA repetition dataset~\citep{runia2018}. 
The models trained with our non-rigid flow set 
capture better delineation and details of the objects, especially for non-rigid
movements of human body parts (indicated by the changes of colors).

\subsection{Realism test}
In this experiment, the goal is to assess the realism of the optical flow
generated by the ARAP deformation model. The baseline is PwC-Net pre-trained
on the FlyingChairs dataset. We compare PWC-Net 
finetuned using ground truth Sintel flow and ones finetuned with ARAP-generated flow on the Sintel-train images and Sintel movie. The results are presented in Table~\ref{tab:realism}. 
The results from finetuning on ground truth flow are considered the theoretical maximum as GT flow is not always available in reality, while the distribution is more aligned with the test set distribution. The results show that we improve our baseline and achieve results that are closer to the theoretical upper bound. Therefore, ARAP-generated flow appears realistic and captures the data distribution.

\begin{table}[t]
    \centering
    \caption{PWC-Nets finetuned on ARAP-generated flow perform closely to one finetuned on the ground truth flow of Sintel and outperfom the baseline, showing the usefulness and realism of the generated flow.
    }
    \begin{adjustbox}{width=.7\columnwidth,valign=b}
    \begin{tabular}{@{}ccccc@{}}\toprule
 	\multirow{2}{*}{Training dataset} & \multicolumn{2}{c}{Sintel-test} & \multicolumn{2}{c}{Sintel-test occ} \\
 	\cmidrule(r){2-3}\cmidrule(l){4-5}
                         & final & clean & final & clean \\
        \midrule        
        FlyingChairs & 6.97 & 5.61 & 33.58 & 30.61  \\         
        ft Sintel-GT & \bf 6.22 & \bf 5.38 & \bf 30.46 & \bf 29.69 \\
        ft ARAP-SintelM & 6.48 & 5.50 & 30.08 & 32.67 \\
        ft ARAP-Sintel & 6.63 & 5.47 & 33.02 & 31.02 \\
        \bottomrule
    \end{tabular}
    \end{adjustbox}
    \label{tab:realism}
\end{table}

%% file: sections/experiments_table_sota.tex
\begin{table}[t]
    \centering
    \caption{
    Comparison of different models (FNS, PWC, LFN) trained on FC and DMO, evaluated on Sintel benchmark, Human Flow, and KITTI.
    The constant zero flow indicates the displacement statistics of the test sets. 
    The results of DeepFlow is reported by 
    \cite{Revaud2015} and DeepFlow* by \cite{Weinzaepfel2013}.
    For all networks and all evaluations hold that training on non-rigid motion data (DMO) outperforms training on rigid/affine motion (FC).
    }
    \adjustbox{width=\columnwidth}{
    \setlength{\tabcolsep}{4pt} %
    \renewcommand{\arraystretch}{1.2} %
    \begin{tabular}{@{}ccccccccc@{}}\toprule
        \multicolumn{2}{c}{} & \multicolumn{2}{c}{Sintel-test} & \multicolumn{2}{c}{Sintel-test occ} & 
        \multirow{2}{*}{Human flow} &
        \multicolumn{2}{c}{KITTI val} \\
        \cmidrule(r){3-4}
        \cmidrule(r){5-6}
        \cmidrule(r){8-9}
        && final & clean & final & clean && 2012 & 2015 \\
        \midrule
        \multicolumn{2}{c}{Zero flow}      & - & - & - & - & 0.73 & 28.23 & 24.03 \\
        \multicolumn{2}{c}{DeepFlow}      & 6.93 & - & 38.17 & - & - & - & - \\
        \multicolumn{2}{c}{DeepFlow*}      & 7.21 & - & 38.78 & - & - & - & - \\
        \midrule
        \multirow{2}{*}{\rotatebox{90}{FNS}} & FC      & 8.16 & 7.17 & 35.88 & 34.02 & 0.63 & 4.63 & 7.71 \\
        &DMO    & \bf 7.64 & \bf 6.61 & \bf 34.98 & \bf 33.17 & \bf 0.36 & \bf 3.53 & \bf 5.30 \\
        \midrule
        \multirow{2}{*}{\rotatebox{90}{LFN}}& FC    & 7.89 & 6.77 & 38.79 & 37.28 & 0.30 & 2.75 & 7.61\\ 
        &DMO      &  \bf 7.73 & \bf 6.50 & \bf 38.68 & \bf 36.30 & \bf 0.26 & \bf 2.73 & \bf 6.27 \\
        \midrule
        \multirow{2}{*}{\rotatebox{90}{PWC}}& FC    & 6.97 & 5.61 & 33.58 & 30.61 & 0.30 & 2.22 & 5.36 \\ 
        &DMO    &   \bf 6.62 & \bf 5.52 & \bf 31.56 & \bf 30.00  & \bf 0.26 & \bf 1.72 & \bf 3.18 \\
        \midrule
        \multirow{2}{*}{\rotatebox{90}{\ud{RAFT}}}& FC    & 7.69 & \bf 4.48 & 35.11 & \bf 24.99 & \bf 0.21 & 3.55 & 6.17 \\ 
        &DMO    &   \bf 6.15 & 4.86 & \bf 32.75 & 28.39  & 0.25 & \bf 1.45 & \bf 2.58 \\
        \bottomrule
    \end{tabular}
    }
    \label{tab:sota}
\end{table}

%% file: sections/experiments_table_ft.tex
\begin{table}[t]
    \centering
    \caption{Comparison to unsupervised (U), transferred (T) and fine-tuned methods (Ft). 
    PWC trained on DMO outperforms all the transferred methods and compares favourably to fine-tuned methods after pre-training on FC.
    The gap between unsupervised and transferred methods and fine-tuned methods indicate the necessity of annotated optical flow.
    }
    \adjustbox{width=\columnwidth}{
    \setlength{\tabcolsep}{2pt}
    \begin{tabular}{@{}ccccccc@{}}\toprule
 	& \multirow{2}{*}{Training dataset} & &  \multicolumn{2}{c}{Sintel-test} & \multicolumn{2}{c}{Sintel-test occ} \\
 	\cmidrule(r){4-5} \cmidrule(l){6-7}
    && & final & clean & final & clean \\
        \midrule
        MFOF & RoamingImages & T & 8.81 & 7.23 &  39.70 & 36.78\\
        DDFlow & FC & T & 7.40 & 6.18 & 39.94 & 38.05 \\
        \midrule
        SelFlow & SintelM & U & 6.57 & 6.56 & 34.72 & 38.30 \\        
        SelFlow & SintelM $\rightarrow$ KITTI $\rightarrow$ Sintel & Ft &  \bf 4.26 & \bf 3.74  & \bf 22.37 & \bf 22.50 \\
        \midrule
        PWC & FC & T &   6.97 & 5.61 & 33.58 & 30.61  \\         
        PWC & FC $\rightarrow$ Sintel & Ft & 6.22 & 5.38 & 30.46 & 29.69 \\
        \midrule
        PWC & DMO & U/T &  6.62 & 5.52 & 31.56 &  30.00  \\        
        PWC & DMO $\rightarrow$ Sintel & Ft &   \bf 5.86 & \bf  5.26 & \bf  29.09 & \bf 29.75 \\
        \bottomrule
    \end{tabular}
    }%
    \label{tab:unsupervised}
\end{table}    

%% file: sections/experiments_figure_quva.tex
\begin{figure}[t]
    \centering
        \includegraphics[width=\columnwidth, valign=b, ]{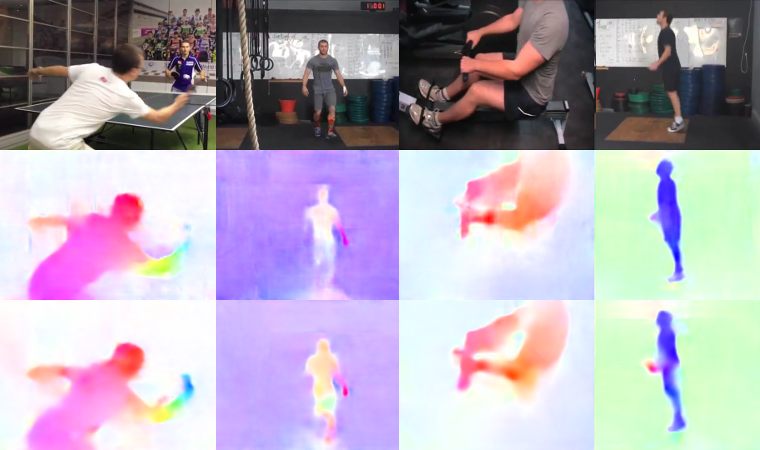}
        \label{fig:QUVA_pwc}
    \caption{Qualitative results on QUVA dataset~\cite{runia2018}
    PWCNet trained on FC (midlle) and DMO (bottom).
    The networks trained using our pipeline capture the non-rigid
    motion of objects in the scenes with higher detail and delineation. (Best viewed in color.)
    }
    \label{fig:quva}
\end{figure}

%% file: sections/conclusion.tex
\section{Conclusion and Discussion}
In this paper, we introduced a pipeline to generate densely annotated optical flow datasets from videos to train supervised deep networks for optical flow.
The method employs a matching process to capture the motion characteristics in videos while varying
objects' textures to increase appearance variations. 
An as-rigid-as-possible (ARAP) image deformation is performed and constrained on the pixel matches to obtain an optical flow field. The flow field is used to warp the first-frame input to create the second frame of the pair for which it is the ground truth flow field, and thus guarantee its correctness.
Extensive experiments are done on the framework to pick the best performance configuration,~\ie using DeepMatching and off-the-shelf Mask-RCNN. The generated dataset tested on several state-of-the-art optical flow prediction architectures show favorable results over the commonly used FlyingChairs for pre-training purposes.

\ud{
The method is propose to generate optical flow data, yet it is also beneficial for analysis purposes by isolating various factors that could affect optical flow learning and prediction.
On the one hand,
it could be seen that optical flow of objects' complicated 3D motion could be learnt with high effectiveness from its simplified 2D deformation.
On the other hand, it provides a new insight that the intermediate components do not need to be free of errors to obtain a dataset that improves performance. The artifacts due to mismatches and uncertainties of background segmentation show favorable for training a network robust to imagery artifacts. This could open a new direction for exploring optical flow datasets and our method can be used as a controlled way to generate optical flow with high variability.
}

There exist many robust and accurate non CNN-based dense optical flow methods, e.g.~\cite{Trinh2019}. Despite working for large and small displacements, scenes with varying textures and under strong illumination changes, 
the goal of such methods and ours is different. The method predicts optical flow from descriptors modelling biological homologous image regions and assumes small neighboring changes in complex scenery. In contrast, our paper aims to generate a dataset guided by motion statistics captured by the matching process with the purpose to train optical flow networks. On the other hand, the theoretical basis for illumination robustness described in the given paper could be considered in a follow-up research to produce more robust matching with reduced artifacts.

\ud{
In the paper, the image deformation as-rigid-as-possible is employed in conjunction with deep matching to capture non-rigid motion characteristics. However, any method that generates optical flow fields can be used as long as the second frame of the pair is created by warping the first-input frame with the generated flow.
Exploring this direction will lead to better understandings of optical flow datasets.
}

{
\small
\noindent \textbf{Acknowledgements:} This work is performed within the TrimBot2020 project funded by the EU Horizon 2020 program No. 688007.
}